\documentclass{article}

\usepackage{microtype}
\usepackage{graphicx}
\usepackage{booktabs} %
\usepackage{multirow}

\usepackage{caption}
\usepackage{physics}
\usepackage{enumerate}

\usepackage{amsmath}
\usepackage{amsfonts}
\usepackage{amsthm}
\usepackage{subfig}

\usepackage{algpseudocode}
\usepackage{varwidth}

\usepackage{color,soul}

\usepackage{hyperref}

\usepackage[accepted]{icml2021}

\icmltitlerunning{Safe Deep Reinforcement Learning for Multi-Agent Systems with Continuous Action Spaces}

\begin{document}

\twocolumn[
\icmltitle{\textbf{Safe Deep Reinforcement Learning for Multi-Agent Systems with Continuous Action Spaces }}

\icmlsetsymbol{equal}{*}

\begin{icmlauthorlist}
\icmlauthor{Ziyad Sheebaelhamd}{equal,to}
\icmlauthor{Konstantinos Zisis}{equal,to}
\icmlauthor{Athina Nisioti}{equal,to}
\icmlauthor{Dimitris Gkouletsos}{equal,to}\\
\icmlauthor{Dario Pavllo}{ed}
\icmlauthor{Jonas Kohler}{ed}
\end{icmlauthorlist}

\icmlaffiliation{to}{ETH Zürich, Switzerland}
\icmlaffiliation{ed}{Department of Computer Science, ETH Zürich, Switzerland}

\icmlcorrespondingauthor{Athina Nisioti}{anisioti@ethz.ch}

\icmlkeywords{Machine Learning, ICML, Reinforcement Learning, Deep RL}

\vskip 0.3in
]

\printAffiliationsAndNotice{\icmlEqualContribution} %

\begin{abstract}
Multi-agent control problems constitute an interesting area of application for deep reinforcement learning models with continuous action spaces. Such real-world applications, however, typically come with critical safety constraints that must not be violated. In order to ensure safety, we enhance the well-known multi-agent deep deterministic policy gradient (MADDPG) framework by adding a safety layer to the deep policy network. %
In particular, we extend the idea of linearizing the single-step transition dynamics, as was done for single-agent systems in Safe DDPG \cite{dalal2018safe}, to multi-agent settings. We additionally propose to circumvent infeasibility problems in the action correction step using soft constraints \cite{kerrigan2000soft}. Results from the theory of exact penalty functions can be used to guarantee constraint satisfaction of the soft constraints under mild assumptions. We empirically find that the soft formulation achieves a dramatic decrease in constraint violations, making safety available even during the learning procedure. 
\end{abstract}

\section{Introduction and Related Work}

In recent years, deep reinforcement learning (Deep RL) with continuous action spaces has received increasing attention in the context of real-world applications such as autonomous driving \cite{driving}, single-~\cite{gu2017deep} and multi robot systems \cite{hu2020voronoi}, as well as data center cooling \cite{cooling}. Contrary to more mature applications of RL such as video games \cite{mnih2015human}, these real-world cases naturally require a set of safety constraints to be fulfilled (e.g.\ in the case of robot arms, avoiding obstacles and self-collisions, or limiting angles). The main caveat of safety in reinforcement learning is that the dynamics of the system are a-priori unknown and hence one does not know which actions are safe ahead of time. Whenever accurate offline simulations or a model of the environment are available, safety can be introduced ex-post by correcting a learned policy for example via shielding \cite{alshiekh2018safe} or via a pre-determined backup controller \cite{wabersich2018linear}. Yet, many real-world applications require safety to be enforced during both learning and deployment, and a model of the environment is not always available.

A growing line of research addresses the problem of safety of the learning process in model-free settings. A traditional approach is reward shaping, where one attempts to encode information on undesirable actions state pairs in the reward function. Unfortunately, this approach comes with the downside that unsafe behavior is discouraged only as long as the relevant trajectories remain stored in the experience replay buffer. In \cite{lipton2018combating}, the authors propose Intrinsic Fear, a framework that mitigates this issue by training a neural network to identify unsafe states, which is then used in shaping the reward function. Although this approach alleviates the problem of periodically revisiting unsafe states, it is still required to visit those states to gather enough information to avoid them.

Another family of approaches focuses on safety for discrete state/action spaces and thus studies the problem through the lens of finite Constrained Markov Decision Processes (CMDPs) \cite{altman1998constrained}. Along those lines, multiple approaches have been proposed. For example in \cite{efroni2020explorationexploitation} the authors propose a framework which focuses on learning the underlying CMDP based purely on logged historical data. A common limitation to such approaches is that it is hard to generalize to continuous action spaces, although there exists some work on that direction, as in \cite{chow2019lyapunovbased} where the authors leverage Lyapunov functions to handle constraints in continuous settings.

For safe control in physical systems, where actions have relatively short-term consequences, \cite{dalal2018safe} propose an off-policy Deep RL method that efficiently exploits single-step transition data to estimate the safety of state-action pairs, thereby successfully eliminating the need of behavior-policy knowledge of traditional off-policy approaches to safe exploration. In particular, the authors directly add a safety layer to a single agent's policy that projects unsafe actions onto the safe domain using a linear approximation of the constraint function, which allows the safety layer to be casted  as a quadratic program. This approximation arises from a first-order Taylor approximation of the constraints in the action space, whose sensitivity is parameterized by a neural network, which was pre-trained on logged historical data.

In this work, we propose a multi-agent extension of the approach presented in \cite{dalal2018safe}. We base our method on the MADDPG framework \cite{lowe2017multi} and aim at preserving safety for all agents during the whole training procedure. Thereby, we drop the conservative assumptions made in \cite{dalal2018safe}, that the optimization problem that corrects unsafe actions only has one constraint active at a time and is thus always feasible. In real world problems, the optimization formulation proposed has no guarantees to be recursively feasible\footnote{Even when following valid actions, agents can end up in states from which safety is no longer recoverable.} and in multi-agent coordination problems where agents impose constraints on one another, one always has more than one constraint active due to the natural symmetry. Instead, we propose to use a specific soft constrained formulation of the problem that addresses the lack of recursive feasibility guarantees in the hard constrained formulation. This enhances safety significantly in practical situations and is general enough to capture the complicated dynamics of multi-agent problems. %
This approach supersedes the need for a backup policy (as in e.g. \cite{Shielding} and \cite{safe_multi_robot}) because the optimizer is allowed to loosen the constraints by a penalized margin as proposed in \cite{kerrigan2000soft}. Thus, our approach does not guarantee zero constraint violations in all situations examined, but by tightening the constraints by a tolerance, one could achieve almost safe behavior during training and in fact, we observe only very rare violations in an extensive set of simulations (Section \ref{sec:results}).

In summary, our contribution lies in extending the approach proposed in \cite{dalal2018safe} (Safe DDPG) to a multi-agent setting, while efficiently circumventing infeasibility problems by reformulating the quadratic safety program in a soft-constrained manner.

\section{Models and Methods}

\subsection{Problem Formulation}
We consider a discrete-time, finite dimensional, decentralized, non-cooperative multi-agent system with $N$ agents, continuous state spaces $\mathcal{X}_{i}$ such that the state $ {x}_{i} \in \mathcal{X}_{i} \subseteq \mathbb{R}^{d} $, continuous action spaces $ \mathcal{A}_i$ such that the action $ {a}_{i} \in \mathcal{A}_i \subseteq \mathbb{R}^{m}$ and a reward function for each agent $R_{i}, \quad \forall i \in\{1,...,N\} $. For clarity, we compactly denote $\boldsymbol{x} = (x_1, \ldots , x_N), \: \boldsymbol{a} = (a_1, \ldots , a_N) $ and $\boldsymbol{R}= (R_1, \ldots, R_N)$. The superscript $t$ is used to denote the time index. In addition, we define a set of $K$ constraints as mappings of the form $c_{j}(\boldsymbol{x}) \quad \forall j \in \{1,...,K\}$, meaning that each constraint may depend on the state of more than one agent. Finally, we define a policy $\pi_{i}$ to be a function mapping the state of agent $i$ to its local action. In the scope of this work, we consider deterministic policies parameterized by $\boldsymbol{\theta} = \qty(\theta_1, ..., \theta_N)$, and thus use the notation $\pi_{\theta_i}$. In this context, we examine the problem of safe exploration in a constrained Markov Game (CMG) and therefore we aim to solve the following optimization problem for each agent:

\begin{equation}\label{eq:init_prob}
    \begin{array}{l}
    \underset{\theta_i} \max \: \mathbb{E} \big[ \sum_{t=0}^{\infty} \gamma^{t} R_{i}\left(x_{i}^{t}, \pi_{\theta_i}(x^{t}_{i})\right)\big], \quad \forall i \\
    \text { s.t. } \quad c_{j}\left(\boldsymbol{x}^t\right) \leq C_{j}, \quad \forall j
    \end{array}
\end{equation}

where $\gamma \in (0,1)$ denotes the discount factor. The above expectation is taken with respect to all agents future action/state pairs - quantities that depend on the policy of each agent - which gives rise to a well-known problem in multi-agent settings, namely the non-stationarity of the environment from the point of view of an individual agent \cite{lowe2017multi}. Alongside the constraint dependence on multiple agents, this is what constitutes the prime difficulty in guaranteeing safety in decentralized multi-agent environments.

It is worth stating that our goal is not only to enhance safety in the solution of the RL algorithm, but also do so during the training procedure. This is relevant for  applications such as self-driving cars and self-flying drones, which require safety in their whole operating period but are too complex to be simulated off-line with high accuracy.

\subsection{Safety Signal Model}
Following \cite{dalal2018safe}, we make a first order approximation of the constraint function in \eqref{eq:init_prob} with respect to action~$\boldsymbol{a}$
\begin{equation}\label{eq:lin_con}
c_{j}\left(\boldsymbol{x}^{\prime}\right) = \hat{c}_{j}(\boldsymbol{x}, \boldsymbol{a}) \approx c_{j}(\boldsymbol{x})+g\left(\boldsymbol{x} ; w_{j}\right)^{\top} \boldsymbol{a},
\end{equation}
where $ \boldsymbol{x} ^{\prime}$ denotes the state that followed $ \boldsymbol{x} $ after applying action $\boldsymbol{a}$ and the function $g$ represents a neural network with input $\boldsymbol{x}$, output of the same dimension as the action $\boldsymbol{a}$ and weights $w_{j}$. This network efficiently learns the constraints' sensitivity to the applied actions given features of the current state based on a set of single-step
transition data $\mathcal{D}=\left\{\left(\boldsymbol{x}^k, \boldsymbol{a}^{k}, \boldsymbol{x}^{k \prime}\right)\right\}$. 

In our experiments, we generate $\mathcal{D}$ by initializing agents with a random state and choosing actions according to a sufficiently exploratory (random) policy for multiple episodes. With the generated data, the sensitivity network can be trained by specifying the loss function for each constraint as
\begin{equation} 
\mathcal{L}(w_j) =  \sum_{\left(\boldsymbol{x}, \boldsymbol{a}, \boldsymbol{x}^{\prime}\right) \in D}\left(c_{j}\left(\boldsymbol{x}^{\prime}\right)-\left(c_{j}(\boldsymbol{x})+g\left(\boldsymbol{x} ; w_{j}\right)^{\top} \boldsymbol{a}\right)\right)^{2} 
\end{equation}
where each constraints' sensitivity will be trained separately.

\begin{figure*}[t]
	\centering
	\includegraphics[width=0.6\linewidth]{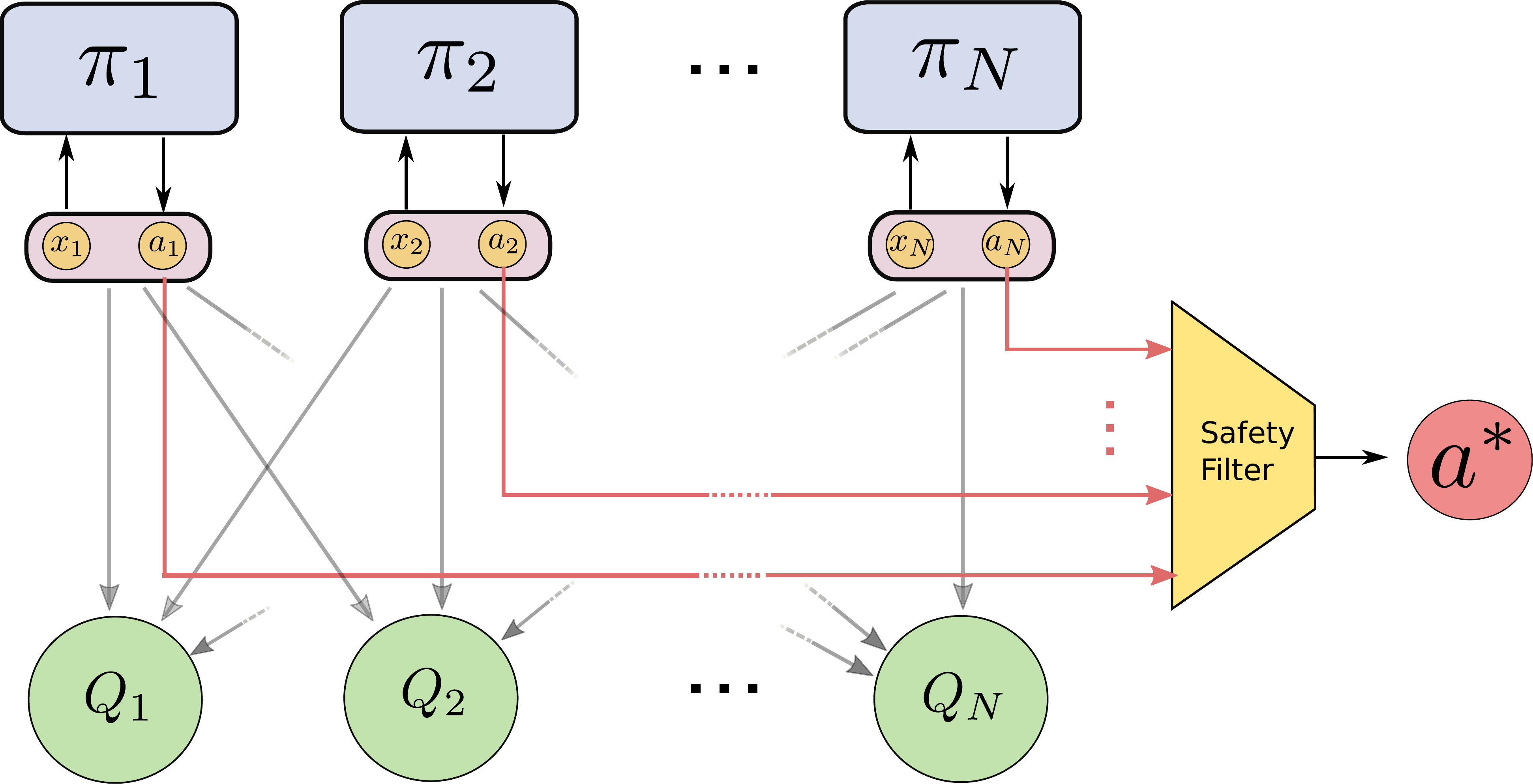}
	\caption{\textbf{An illustration of the safety layer used in combination with the MADDPG networks in order to apply the safe projection of the optimal action}. The individual states of all agents $x_i$ are fed into their corresponding policy networks, outputting the evaluation of their current policies at those states which are then concatenated into a single vector. Finally, a convex quadratic optimization problem is solved in order to produce the optimal safe action $\boldsymbol{a}^*$.}
	\label{fig:pipeline}
	\end{figure*}

\subsection{Safety Layer Optimization}
Given the one-step safety signals introduced in \eqref{eq:lin_con}, we augment the policy networks by introducing an additional centralized safety layer, which enhances safety by solving
\begin{equation}\label{eq:opt2}
\begin{array}{l}
\underset{\boldsymbol{a}}{\arg \min } \left\|\boldsymbol{a}-\Pi(\boldsymbol{x})\right\|^{2}_2 \\
\text { s.t. }  c_{j}(\boldsymbol{x})+g\left(\boldsymbol{x} ; w_{j}\right)^{\top} \boldsymbol{a} \leq C_{j} \quad \forall j = \{1, \ldots, K\} \text {, }
\end{array}
\end{equation}
where $\Pi(\boldsymbol{x})$ denotes the concatenation of all local agents' policies, i.e. $\Pi(\boldsymbol{x}) = \qty( \pi_{\theta_1}(x_1),..., \pi_{\theta_N}(x_N))$.
This constitutes a quadratic program which computes the (minimum distance) projection of the actions proposed by each of the policy networks $\pi_{\theta_{i}}(x_{i})$ onto the \textit{linearized} safety set. Figure \ref{fig:pipeline} illustrates the whole pipeline for computing an action from a given state. 

Due to the strong convexity of the resulting optimization problem, there exists a global unique minimizer to the problem whenever the feasible set is non-empty. In contrast to \cite{dalal2018safe}, where recursive feasibility was assumed and therefore a closed form solution using the Lagrangian multipliers was derived, we used a numerical QP-solver to defer from making this rather strong assumption on the existence of the solution, which is not guaranteed for dynamical systems. %

Due to the generality of the formulation, it is possible that there exists no recoverable action that can guarantee the agents to be taken to a safe state although the previous iteration of the optimization was indeed feasible. The reason for that is that we assume a limited control authority, which further must respect the dynamics of the underlying system. To take this into account without running into infeasibility problems where the agents would require a backup policy to exit unrecoverable states, we propose a soft constrained formulation, whose solution is equivalent to the original formulation whenever \eqref{eq:opt2} is feasible. Otherwise, the optimizer is allowed to loosen the constraints by a penalized margin as proposed in \cite{kerrigan2000soft}. We thus reformulate \eqref{eq:opt2} as follows
 
\begin{equation}\label{eq:opt3}
\begin{aligned}
& (\boldsymbol{a}^{*}, \boldsymbol{\epsilon}^{*})= \underset{\boldsymbol{a}, \boldsymbol{\epsilon}}{\arg \min } \left\|\boldsymbol{a}-\Pi(\boldsymbol{x})\right\|_2^{2} +
\rho \left\|\boldsymbol{\epsilon} \right\|_1\\
& \text { s.t. } g\left(\boldsymbol{x} ; w_{j}\right)^{\top} \boldsymbol{a} \leq C_{j} -c_{j}(\boldsymbol{x}) + \epsilon_i  \\
\quad & \quad \quad \quad \epsilon_j \geq 0 \text { . }
\quad \forall j = \{1, \ldots, K\}
\end{aligned}
\end{equation}
where $\boldsymbol{\epsilon} = \left( \epsilon_1, \ldots, \epsilon_K \right)$
are the slack variables and $\rho$ is the constraint violation penalty weight. We pick $\rho >  \left\|\lambda^* \right\|_{\infty}$ where $\lambda^*$ is the optimal Lagrange 
multiplier for the original problem formulation in \eqref{eq:opt2}, which 
guarantees that the soft-constrained problem yields
equivalent solutions whenever (\ref{eq:opt2}) is feasible (see \cite{kerrigan2000soft}). 
Since exactly quantifying the optimal Lagrange multiplier is time-consuming, we assign a large value of $\rho$
by inspection. It is important to mention that the reformulation in (\ref{eq:opt3}) 
still constitutes a quadratic program when extending the optimization 
vector into $(\boldsymbol{a}, \boldsymbol{\epsilon})$ and using an epigraph formulation \cite{rockafellar2015convex}.
Notably, this formulation does not necessarily guarantee zero constraint violations. However, we observe empirically that violations remain very small, when setting a rather high penalty value $\rho$ (see Figure \ref{fig:collisions_training}).

\subsection{Multi-Agent Deep Deterministic Policy Gradient Algorithm (MADDPG)}
For training Deep RL agents in continuous action spaces, the use of policy gradient algorithms, in which the agent's policy is directly parameterized by a neural network, is particularly well suited as it avoids explicit maximization over continuous actions which is intractable. We thus opt for the Multi-Agent Deep Deterministic Policy Gradient (MADDPG) algorithm \cite{env1} which is a multi-agent generalization of the well-known DDPG methods, originally proposed in \cite{DDPG}.

The MADDPG algorithm is in essence a multi-agent variation of the Actor-Critic architecture, where the problem of the environment's non-stationarity is addressed by utilizing a series of centralized Q-networks which approximate the agents' respective optimal Q-value functions using full state and action information. This unavoidably enforces information exchange during training time, which is sometimes referenced as ``centralized training". On the other hand, the actors employ a policy gradient scheme, where each policy network has access to agent specific information only. Once the policy networks converge, only local observations are required to compute each agent's actions, thus allowing decentralized execution.

For stability purposes, MADDPG incorporates ideas from Deep Q Networks, originally introduced in \cite{mnih2013playing}. Specifically, a replay buffer $\mathcal{R}$ stores historical tuples $\qty(\boldsymbol{x}, \boldsymbol{a}, \boldsymbol{R}, \boldsymbol{x}^\prime)$, which can be used for off-policy learning and also for breaking the temporal correlation between samples. Furthermore, for each actor and critic network, additional target networks are used to enhance stability of the learning process. We denote as $Q_i^\pi(\boldsymbol{x}, \boldsymbol{a}; \beta_{i})$ the critic network, parameterized by $\beta_{i}$, and as $\pi_i(x_i; \theta_i)$ the actor network for agent $i$, parameterized by $\theta_i$. As for the target networks we denote them as ${\hat{Q}_i^{\pi }} (\boldsymbol{x}, \boldsymbol{a}; \hat{\beta}_{i})$ and $\hat{\pi}_i(x_i; \hat{\theta}_i)$ respectively. %
Finally, we use $\tau$ to denote the convex combination factor for updating the target networks.

\begin{algorithm}[h]
\begin{algorithmic}[1]
\State Initialize random weights $\beta_i$ and $\theta_i$ of critic network $Q_i^\pi(\boldsymbol{x}, \boldsymbol{a}; \beta_{i})$ and actor network $\pi_i(x_i; \theta_i)$
\State Initialize weights $\hat{\beta}_i \leftarrow \beta_i $ and $\hat{\theta}_i \leftarrow \theta_i$ of target networks ${\hat{Q}_i^\pi}$ and $\hat{\pi}_i$

\State Initialize a replay buffer $\mathcal{R}=\{\}$

\For{$episode = 1$ to $m $}
    \State Randomly initialize the state $\boldsymbol{x}^1$
    \For{$t = 1$ to $T $}
        \State Pick action $a_i^{t}=\pi_i \qty(x_{i}^t ; \theta_{i})$ for each agent
        \State Concatenate actions into $\boldsymbol{a}^t = \qty(a_1^t,..., a_N^t)$
        \State Project $\boldsymbol{a}^t$ to the safety set by solving \eqref{eq:opt3}
        \State Inject exploratory noise $n$ %
        \State Apply $\boldsymbol{a}^t$, obtain reward $\boldsymbol{R}^t$ and next state ${\boldsymbol{x}^{t}}^\prime$
        \State Store transition $\left(\boldsymbol{x}^t, \boldsymbol{a}^{t}, \boldsymbol{R}^{t},{\boldsymbol{x}^{t}}^\prime \right)$ in  $\mathcal{R}$
        \State \begin{varwidth}[t]{\linewidth}Sample mini-batch of $\lambda$ transitions from  $\mathcal{R}$,\\ indexed by $k$\end{varwidth}\vspace{2mm}
        \State Compute $\boldsymbol{\tilde{a}} = \qty( \hat{\pi}_1 ( x_{1}^{\prime} ; \hat{\theta}_1 ),..., 
        \hat{\pi}_N (x_{N}^{\prime} ; \hat{\theta}_N ))$
        \State Set $z_{i}^k = R_{i}^k +\gamma {\hat{Q}_i^\pi} \left({\boldsymbol{x}^k}^\prime,  \boldsymbol{\tilde{a}} ; \hat{\beta}_i \right)$

        \State \begin{varwidth}[t]{\linewidth}Update each critic network by minimizing\\$L_i = \frac{1}{\lambda} \sum_{k}\left(z_{i}^k - Q_i^\pi \left(\boldsymbol{x}_{k}, \boldsymbol{\tilde{a}} ; \beta_i \right)\right)^{2}$\end{varwidth}\vspace{2mm}

        \State \begin{varwidth}[t]{\linewidth}Update actor policy of each actor: $\nabla_{\theta_{i}} J_i \approx$\\
         $
             \frac{ \sum_{k} \nabla_{a_i} Q_i^\pi \left(\boldsymbol{x}^{k}, \boldsymbol{a}^{k} ; \beta_i \right)|_{ a_i=\pi_i \left(x_{i}^{k}\right)} \nabla_{\theta_i} \pi_i\left(x_i^{k} ; \theta_i \right)|_{x_{i}^{k} }}{\lambda}
         $\end{varwidth}\vspace{2mm}
         \State \begin{varwidth}[t]{\linewidth}Update target networks for each agent\\
         $
             \hat{\beta}_i \leftarrow \tau \beta_i + 
             (1-\tau) \hat{\beta}_i \\ \hat{\theta}_i \leftarrow \tau \theta_i+(1-\tau) \hat{\theta}_i
         $\end{varwidth}\vspace{2mm}
\EndFor
\EndFor
\end{algorithmic}
\caption{\label{alg:ddpg} Safe MADDPG Algorithm}
\end{algorithm}
\subsection{Implementation Details}
\label{subsec:impl_details}
In order to assess the performance of our proposed method we conducted experiments using the multi-agent particle environment, which was previously studied in \cite{env1} and \cite{mordatch2018emergence}. In this environment, a fixed number of agents are moving collaboratively in a 2-D grid trying to reach specific target positions. In our experiments, we used three agents that are constrained to avoid collisions among them. Each agent's state $x_i$ is composed of a vector in $\mathbb{R}^{10}$, containing its position and velocity, the relative distances to the other agents and the target landmark location. Moreover, the actions $a_i$ are defined as vectors in $\mathbb{R}^{2}$ containing the acceleration on the two axes. 

\begin{figure}[t]
    \centering
	\includegraphics[trim={0cm 0 0cm 0},clip, width=0.7\linewidth]{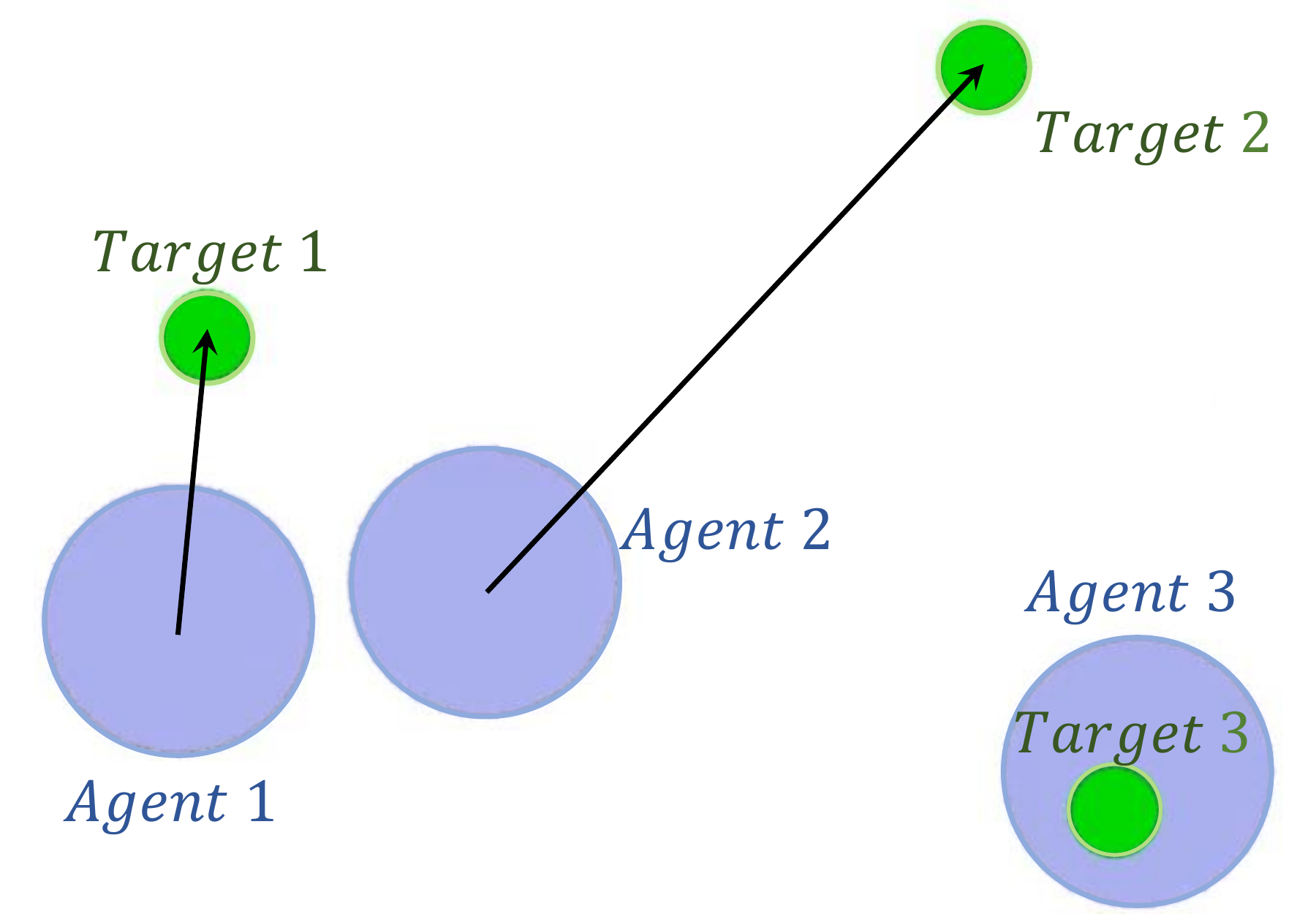}
    \caption{\textbf{A snapshot of the multi-agent particle environment used in the described simulations.} Blue color is used to capture the agents, whereas green color refers to the target positions. In our experiments we have 3 agents trying to move to a specific target without colliding with each other. }
    \label{fig:environment}
\end{figure}

The reward assigned to each agent is proportional to the negative $l_1$ distance of the agent from its corresponding target and furthermore, collisions are being penalized. The agents receive individual rewards based on their respective performance. For the safety layer pre-training, we train a simple fully connected ReLU network $g(\boldsymbol{x}, w_{i})$ with a 10-neuron  hidden layer  for each of the six existing constraints (two possible collisions for each agent), on the randomly produced dataset $\mathcal{D}$. Note that, due to the pairwise symmetry of the constraints used in the experiment (A colliding with B implies B colliding with A), we could in principle simplify the network design, but for generality we decided to consider them as independent constraints. Based on our empirical results (Section \ref{sec:results}), we found it unnecessary to increase the complexity of the network. We train the model using the popular Adam optimizer \cite{kingma2015adam} with a batch size of 256 samples. %
For solving the QP Problem, we adopted the \texttt{qpsolvers} library, which employs a dual active set algorithm originally proposed in \cite{qpsolver}. We further used a value of $\rho = 1000$ in the cost function of the soft formulation. For the MADDPG algorithm implementation, we used three pairs of fully connected actor-critic networks. These networks are composed of two hidden layers with 100 and 500 neurons respectively. The choice for all activation functions is ReLU except for the output layer of the actor networks, where \texttt{tanh} was used to compress the actions in the [-1,1] range and represent the agents' limited control authority. The convex combination rate $\tau$, used when updating the target networks, was set to $0.01$.

To evaluate the algorithm's robustness and its capability of coming up with an implicit recovery policy, we conduct two case studies:\\

\begin{enumerate}%
    \item [\textbf{(ED)}] Inject an \textbf{E}xogenous uniform \textbf{D}isturbance after each step of the environment, which resembles a very common scenario in real life deployment where environment mismatch could lead to such a behaviour.
    \item [\textbf{(UI)}]  Allow the environment to be \textbf{U}nsafely \textbf{I}nitialized, which can also occur in practice. 
\end{enumerate}
\vspace{-20pt}

\section{Results}
\label{sec:results}
We assess the performance of the proposed algorithm on three metrics: average reward, number of collisions during training, and number of collisions during testing (i.e.\ after training has converged). An infeasible 
occurrence appears in case the hard-constrained QP in (\ref{eq:opt2}) fails to determine a solution that satisfies the imposed constraints.

\begin{figure}[t]
    \centering
	\includegraphics[trim={0cm 0 1cm 1cm scale=1.0},clip, width=0.9\linewidth]{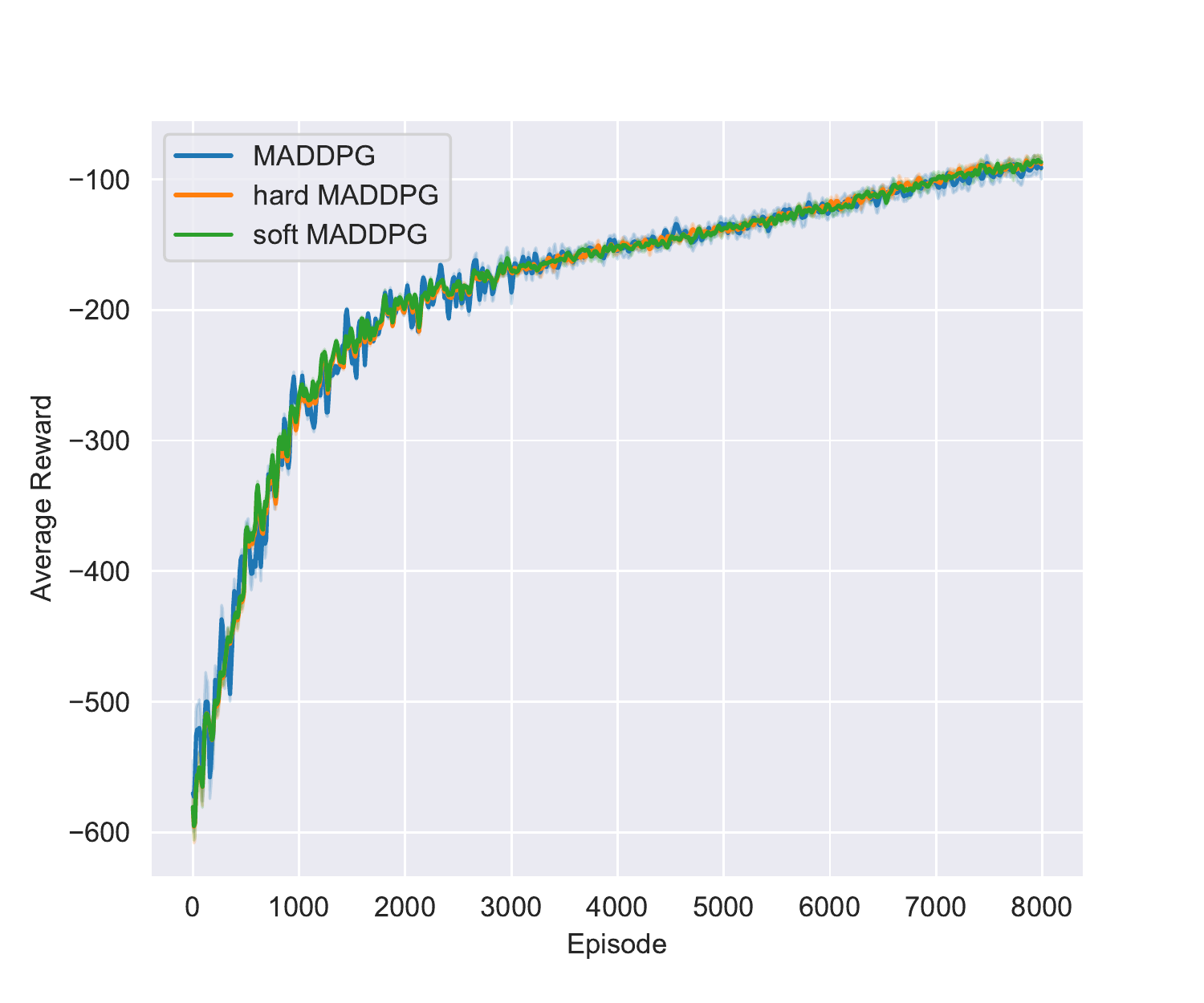}
    \caption{\textbf{An illustration of the rewards achieved during training. ((ED) case)} We observe that all three different procedures have similar reward convergence behaviour, which is not affected by constraints.}
    \label{fig:rewards}
\end{figure}

\begin{figure}[h!]
    \centering
    \includegraphics[trim={0cm 0 1cm 1cm scale=0.9},clip, width=0.9\linewidth]{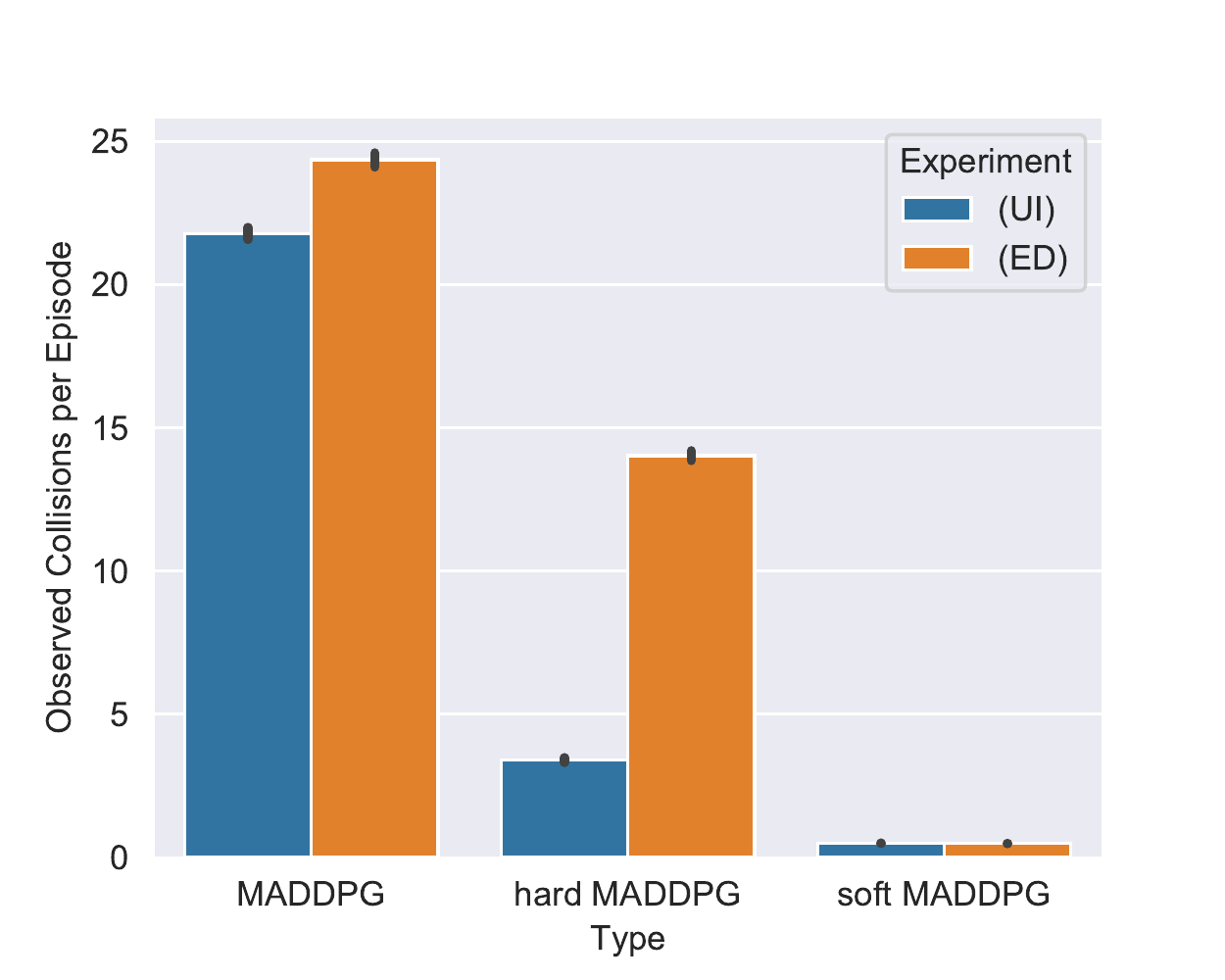}
    \caption{\textbf{An illustration of the average collisions per episode achieved during training for each of the 3 type of agents for both experiments.}} %
    \label{fig:boxplot_training}
\end{figure}

\begin{figure*}[t!] 
    \centering
    \subfloat[Exogenous Disturbances]{%
        \includegraphics[width=0.45\linewidth]{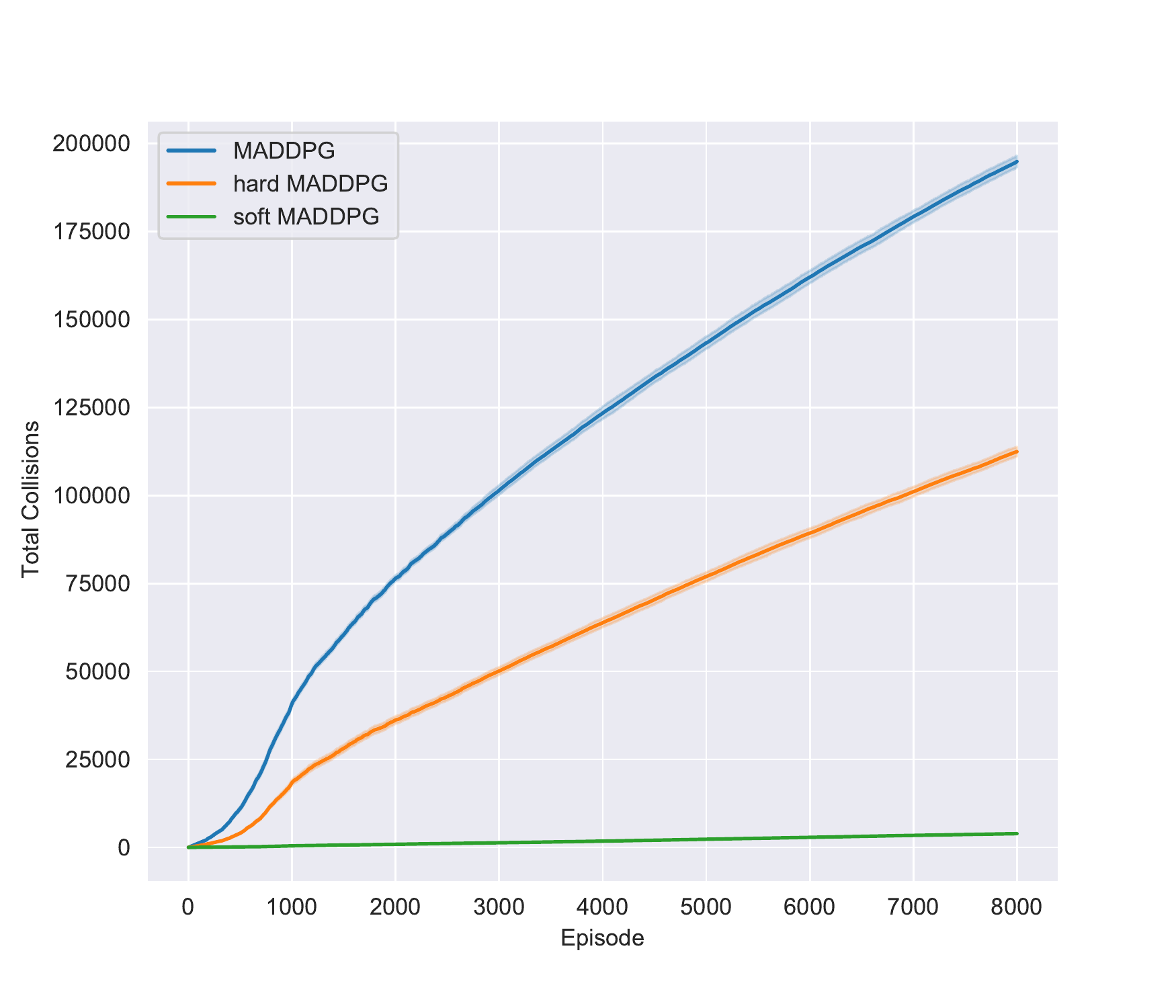}%
        }%
    \hfill%
    \subfloat[Unsafe Initialization]{%
        \includegraphics[width=0.45\linewidth]{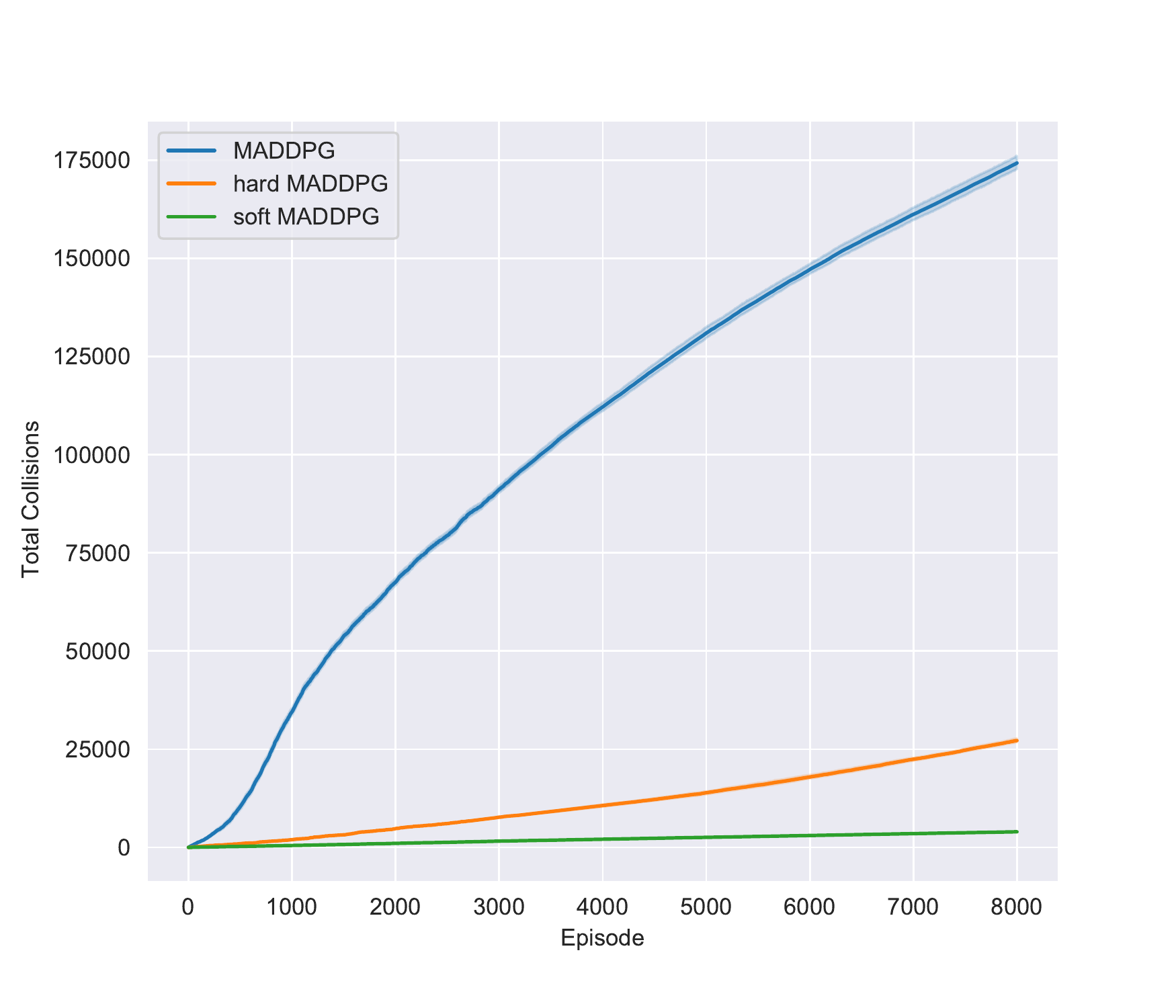}%
        }%
    \caption{ \textbf{An illustration of the cumulative number of collisions achieved during training for the two experiments.} As expected, during the unconstrained MADDPG training, a massive number of collisions is observed, whereas using hard MADDPG, collisions decreased, however, not as significant of a decrease compared to the soft-formulated agent.}
    \label{fig:collisions_training}
\end{figure*}

\begin{figure*}[h!] 
    \centering
    \subfloat[Exogenous Disturbances]{%
        \includegraphics[width=0.45\linewidth]{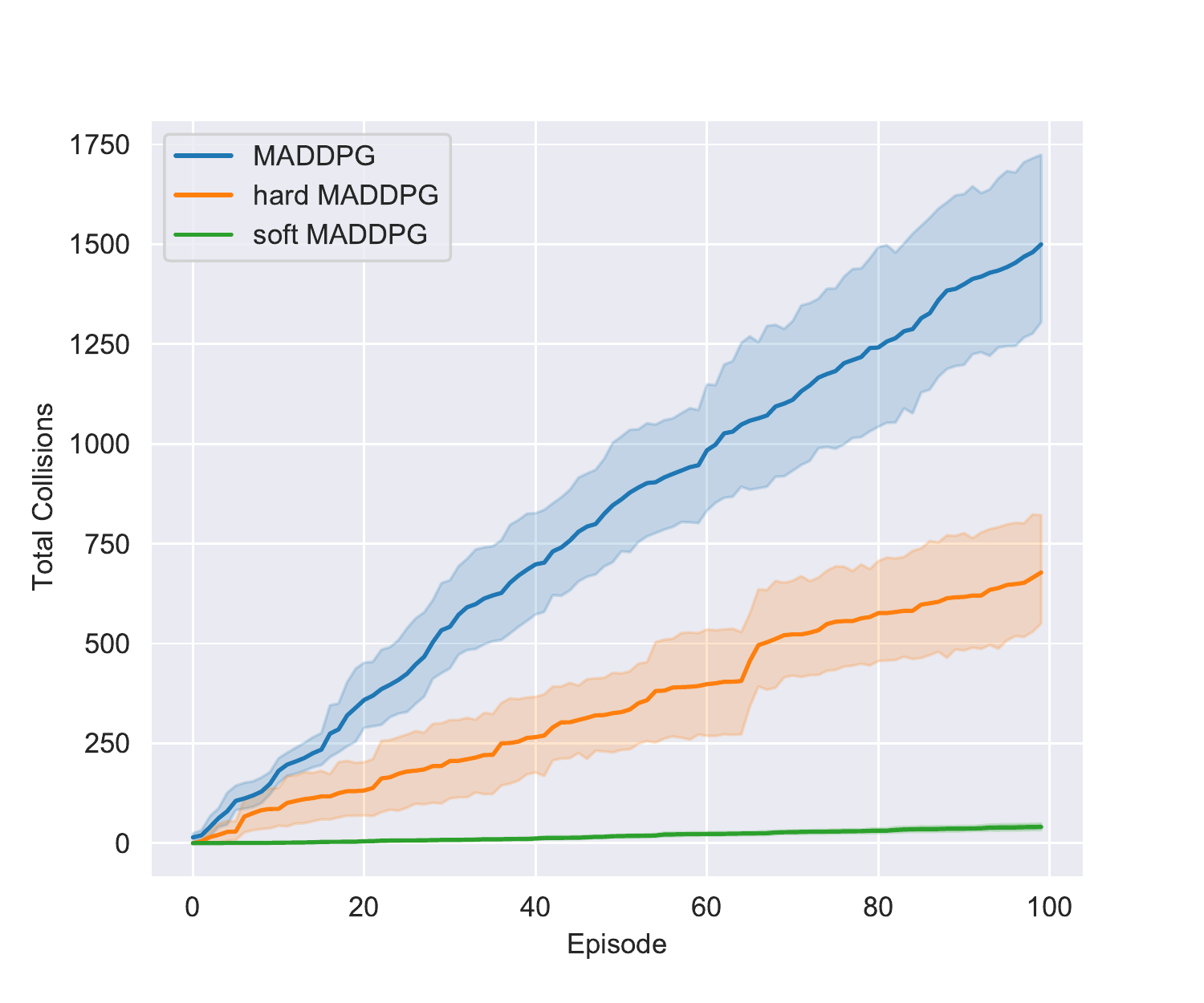}%
        }%
    \hfill%
    \subfloat[Unsafe Initialization]{%
        \includegraphics[width=0.45\linewidth]{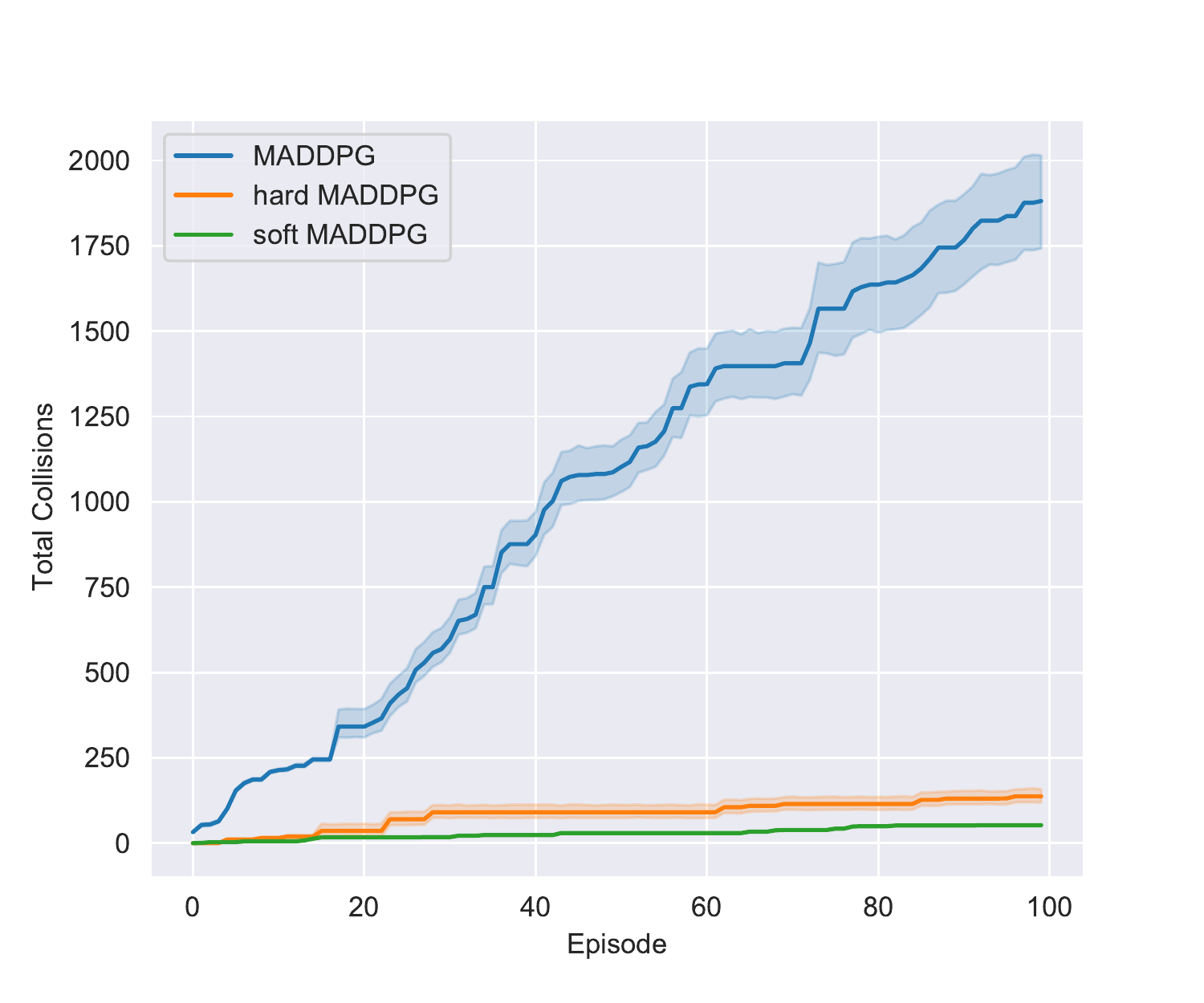}%
        }%
    \caption{ \textbf{An illustration of the cumulative number of collisions during the test simulations performed. } It is crucial to observe that the soft variation shows- also during testing- the smallest number of collisions.}
    \label{fig:collisions_testing}
\end{figure*}

We benchmark a total of three different Deep RL strategies:\vspace{-1.0em}
\begin{itemize}
\setlength\itemsep{-0.1em}
    \item  unconstrained MADDPG,
    \item  hard-constrained MADDPG (hard MADDPG),
    \item  soft-constrained MADDPG (soft MADDPG).
\end{itemize}
 \vspace{-1.0em} The first approach prioritizes exploration and learning over safety since constraints are not directly imposed, whereas hard MADDPG takes into account safe operation by imposing hard state constraints. Finally, soft MADDPG, as presented in Algorithm \ref{alg:ddpg}, imposes a relaxed version of the state constraints while penalizing the amount of slack, following our formulation in \eqref{eq:opt3}.

\begin{table*}[t]
	\centering
	\caption{\textbf{Table illustrating the results of the experiments with the proposed algorithms.} For each setting, we report the mean and corresponding 95\% confidence intervals across  multiple runs. The higher reward is observed in the hard MADDPG, which due to infeasibility problems shows a larger number of collisions during training compared to soft MADDPG, which has the lowest number of collisions. Finally, it is worth observing that the same pattern holds during testing.\\}

	\resizebox{\textwidth}{!}{
		\begin{tabular}{ p{1.8cm} | p{2.4cm} | p{3.6cm} p{5cm} p{5cm}  }
			\toprule 
			\textbf{Experiment}& \textbf{Agent Type} & \textbf{Total Reward \newline (training)} & \textbf{Cumulative number of collisions (training)}& \textbf{Cumulative number of collisions (testing)} \\
			\toprule
			 &MADDPG   & -94.47 \newline 95\%ci: (-108.09, -80.84)   & 174267.44 (baseline = 100\%)  \newline 95\%ci: (172141.24, 176393.64) & 1386.54 \newline 95\%ci: (1263.98, 1509.10)  \vspace{0.2cm} \\
			
			 \hspace{10pt}\textbf{(UI)}&hard MADDPG    & -84.13 \newline 95\%ci: (-95.04, -73.22)   & 27210.44 (15.6\% of baseline)  \newline 95\%ci: (26579.12, 27841.76) \vspace{0.2cm} & 137.44 \newline 95\%ci: (113.02, 161.86)  \\
			&soft MADDPG & \textbf{-136.39} \newline 95\%ci: (-146.79, -125.99)   &  \textbf{3977.0} (2.28\% of baseline)  \newline 95\%ci: (3901.99, 4052.00) & \textbf{52.77} \newline 95\%ci: (52.26, 53.29) \\
			\midrule 
			&MADDPG   & -91.97 \newline 95\%ci: (-103.55, -80.39)   & 194844.0 (baseline = 100\%)  \newline 95\%ci: (192523.44, 197164.55) &1499.77 \newline 95\%ci: (1227.0, 1772.55)  \vspace{0.2cm} \\
			
			\hspace{10pt}\textbf{(ED)}&hard MADDPG    & -89.74 \newline 95\%ci: (-94.53, -84.94)   & 112189.66 (59.57\% of baseline)  \newline 95\%ci: (110405.22, 113974.10) \vspace{0.2cm} & 678.11 \newline 95\%ci: (502.87, 853.34) \\
			&soft MADDPG & \textbf{-86.54} \newline 95\%ci: (-94.22, -78.86)   &  \textbf{3899.11} (2.0\% of baseline)  \newline 95\%ci: (3817.34, 3980.88) & \textbf{40.77} \newline 95\%ci: (30.82, 50.73)  \\
			\bottomrule
		\end{tabular}
	}
	\label{tab:results}
\end{table*}

The duration of each experiment is 8000 episodes and, in order to assess uncertainty, we repeat each experiment 10 times using different initial random seeds.

Under normal operating conditions (safe initialization without disturbances), both the hard and the soft-constrained MADDPG strategies achieve 0 constraint violations in our experiments during the training and the testing phase. However, in order to examine the robustness properties of the aforementioned methods, we evaluate our models under the case studies mentioned in the end of Section \ref{subsec:impl_details}. The outcome of the experiments along with the 95\% confidence intervals are summarized in Table \ref{tab:results}.

In Figure \ref{fig:rewards}, the evolution over episodes of the average reward is depicted for the \textbf{(ED)} case. The average reward is computed as the mean over the  3 agents in a single episode. Interestingly, we observe that all three presented algorithms have a similar 
trend, suggesting that introducing the safety framework (in both hard- and soft- variants) does not negatively affect the ability of the agents to reach their targets. A similar result  holds for the case of unsafe initialization \textbf{(UI)}, so the respective plot is omitted for brevity.

Figure \ref{fig:boxplot_training} shows for each setting, the average number of collisions per episode \emph{during training}, while Figure \ref{fig:collisions_training} presents the evolution of the cumulative number of collisions over 
the training episodes. In particular, soft MADDPG exhibits $97.71 \% $ \textbf{(UI)}, $97.99 \%$ \textbf{(ED)} fewer collisions  compared to the unconstrained MADDPG.  
On the other hand, hard MADDPG only achieves a $ 84.38 \%$ \textbf{(UI)}, $ 42.42 \%$ \textbf{(ED)} reduction in collisions compared to the unconstrained MADDPG, since the infeasibility of the optimization problem does at times not allow the safety filter to intervene and correct the proposed actions. 

To evaluate the impact of the hard-constrained MADDPG, it is essential to investigate the infeasible occurrences, since they represent the critical times when constraints can no longer be satisfied. In our experiments, 20.9\% \textbf{(UI)}, 56.7 \%  \textbf{(ED)} of the episodes are directly related to infeasible conditions. 
This motivates the necessity for a soft-constrained safety layer that maintains feasibility and preserves safety in cases where the hard constrained formulation fails to return a solution. 

Finally, in order to gain a better understanding of the behaviour of our algorithm after convergence, we ran test simulations of 100 episodes for each agent for 10 different initial random seeds. The cumulative number of the collisions for the 2 different settings
is illustrated in Figure \ref{fig:collisions_testing}. It is evident that in both settings, the soft constrained formulation achieves the minimum number of collisions. For visualization purposes, we provide the videos of the test simulation at the following \href{https://www.dropbox.com/sh/9rd5qai96oqmnjs/AABHxhUUOYgsCbo5n3NoG_hva?dl=0}{Video Repository}.

\section{Conclusion}
We proposed an extension of Safe DDPG \cite{dalal2018safe} to multi-agent settings. From a technical perspective, we relaxed some of the conservative assumptions made in the original single-agent work by introducing soft constraints in the optimization objective. This allows us to generalize the approach to settings where more than one constraint is active, which is typically the case for multi-agent environments. Our empirical results suggest that our soft constrained formulation achieves as dramatic decrease in constraint violations during training when exogenous disturbances and unsafe initialization are encountered, while maintaining the ability to explore and hence solve the desired task successfully. %
Although this observation does not necessarily generalize to more complex environments, it motivates the practicality of our algorithm in safety-critical deployment under more conservative constraint tightenings.
Finally, while our preliminary results are encouraging, we believe there is ample room for improvement and further experimentation. As part of future work, we would like to introduce a reward based on the intervention of the safety filter during training, such that we can indirectly propagate the safe behavior to the learnt policies of the agents, which could ultimately eliminate the requirement for using a centralized safety filter during test time. Additionally, we would like to deploy our approach in more complex environments to explore the true potential of our work.

Code can be accessed in the following \href{https://github.com/zisikons/deep-rl}{link}.
\clearpage
\bibliography{references}
\bibliographystyle{icml2021}

\end{document}